\documentclass[10pt,conference]{IEEEtran}
\IEEEoverridecommandlockouts
\usepackage{cite}
\usepackage{amsmath,amssymb,amsfonts}
\usepackage{algorithmic}
\usepackage{graphicx}
\usepackage{textcomp}
\usepackage{xcolor}
\usepackage{float}
\usepackage{subcaption}

\def\BibTeX{{\rm B\kern-.05em{\sc i\kern-.025em b}\kern-.08em
    T\kern-.1667em\lower.7ex\hbox{E}\kern-.125emX}}
\begin{document}

\title{A Novel Statistical Measure for Out-of-Distribution Detection in Data Quality Assurance

}

%\author{\IEEEauthorblockN{1\textsuperscript{st} Tinghui Ouyang}
%\IEEEauthorblockA{\textit{Information and Society Research Division} \\
%\textit{National Institute of Informatics (NII)}\\
%Tokyo, Japan \\
%thouyang@nii.ac.jp}
%\and
%\IEEEauthorblockN{2\textsuperscript{nd} Isao Echizen}
%\IEEEauthorblockA{\textit{Information and Society Research Division} \\
%\textit{National Institute of Informatics (NII)}\\
%Tokyo, Japan \\
%iechizen@nii.ac.jp}
%\and
%\IEEEauthorblockN{3\textsuperscript{rd} Yoshiki Seo}
%\IEEEauthorblockA{\textit{Digital Architecture Research Center} \\
%\textit{National Institute of Advanced Industrial Science and Technology (AIST)}\\
%Tokyo, Japan \\
%y.seo@aist.go.jp}
%}

\author{
Tinghui Ouyang$^{1}$, Isao Echizen$^{1}$, and Yoshiki Seo$^{2}$ \\
\small{\textit{$^{1}$National Institute of Informatics, Japan}}\\
\small{\textit{$^{2}$Digital Architecture Research Center, National Institute of Advanced Industrial Science and Technology, Japan}} \\
\small{E-mail: \{thouyang, iechizen\}@nii.ac.jp; y.seo@aist.go.jp}
}

\maketitle

\begin{abstract}
Data outside the problem domain poses significant threats to the security of AI-based intelligent systems. Aiming to investigate the data domain and out-of-distribution (OOD) data in AI  quality management (AIQM) study, this paper proposes to use deep learning techniques for feature representation and develop a novel statistical measure for OOD detection. First, to extract low-dimensional representative features distinguishing normal and OOD data, the proposed research combines the deep auto-encoder (AE) architecture and neuron activation status for feature engineering. Then, using local conditional probability (LCP) in data reconstruction, a novel and superior statistical measure is developed to calculate the score of OOD detection. Experiments and evaluations are conducted on image benchmark datasets and an industrial dataset. Through comparative analysis with other common statistical measures in OOD detection, the proposed research is validated as feasible and effective in OOD and AIQM studies. 
\end{abstract}


%%%%%%%%% BODY TEXT
\section{Introduction}
In AI quality management (AIQM) \cite{b1}, a reliable AI system is expected to have the ability to make accurate decisions on both data within the problem domain and unknown examples, e.g., to reject data out of the defined problem domain. A well-known example is the postcode recognition system, designed to recognize images of handwriting digital numbers of 0-9. However, things often don't go as planned in the practical operation scenario. An alphabet image is likely fed to the given system for reasons but predicted to a class of 0-9. It is easily understood that the prediction mismatches the realistic expectation because of the absence of a sound problem domain analysis in AIQM. When a similar issue happens in some safety-critical scenarios, like autonomous driving \cite{b3, b4}, no warning or hand-over of driving control may cause serious accidents when facing unusual scenes. Therefore, for the consideration of safety and security, it is vital to investigate the problem domain and study data quality assurance in AIQM, especially detecting the usual data outside the problem domain.

According to the knowledge of general AIs, especially classification or recognition AI systems, their problem domains are usually designed based on closed-world assumptions \cite{b5}. Their training and testing data are also assumed to locate in the same problem domain. These problem domains are usually expressed by data distribution in the perspective of data quality analysis, while the unusual examples are regarded as data of out-of-distribution (OOD) \cite{b6}. Therefore, the easy solution to warning dangerous scenarios in AIQM is effectively detecting OOD data in advance. It is known that the problem of OOD detection is a typical task in AI quality research and is deeply related to anomaly detection (AD) \cite{b7}, novelty detection (ND) \cite{b8}, and open set recognition (OSR) \cite{b9}. Its general idea is to use data distribution or related structural characteristics (e.g., distance, density) for separating normal and outlier data, so the straightforward methods are unsupervised learning-based models. There are many OOD detection methods using different types of structural characteristics reported in the literature. For example, the first type uses probability or density, named probabilistic methods \cite{b10}. These methods leverage the probability density function of a dataset $X$ under a given model parameter $\Theta$, then determine data points having the smallest likelihood $P(X|\Theta)$ as outliers, e.g., the Gaussian Mixture Model (GMM) \cite{b11} fitting a number of Gaussian distributions for outlier detection. Probabilistic principal component analysis (PPCA) \cite{b13} and least-squares anomaly detection (LSA) \cite{b14} are other probabilistic OOD methods. The second type is density-based method. For example, kernel density (KD) estimators \cite{b12} approximate the density function of the dataset via kernel function calculation and distinguish outliers by low density in OOD detection. The third type is based on distance, using the idea that outliers are distant from the distribution of normal data. For instance, Mahalanobis distance (MD) is used to detect anomalies under the assumption of Gaussian-shaped data distribution \cite{b15}. Moreover, there are also some other metrics used in OOD detection methods, e.g., the local outlier factor (LOF) \cite{b16} and $k$NN \cite{b17} leveraging the neighbors' information, a method using the information-theoretic measure (i.e., Kullback-Leibler (KL) divergence \cite{b18}) in OOD investigation.

It is generally straightforward to use these statistical measures to detect OOD data directly in the original data space and easily understood to achieve a good performance on simple data. However, facing complex and high-dimensional datasets, e.g., the image data, direct usage of these measures usually fail to effectively detect OOD because of the bad computational scalability and the curse of dimensionality. Therefore, suitable feature engineering is required before applying these statistical measures in the OOD study. Concerning the above issues, this paper proposes a novel OOD method with consideration of both effective feature engineering and a useful statistical measure. The concrete novelties and contributions of the proposed method are summarized as follows:

1) A feature extraction method based on autoencoder (AE) \cite{b19} and neuron activation status is proposed. It is commonly known that deep learning (DL) has excellent feature learning ability and wildly succeeds in complex data, e.g., image-based data processing. Meanwhile, as a typical DL algorithm for feature learning, deep AEs attempt to learn low-dimensional salient features to reproduce the original data. Therefore, to preserve image data's distribution information as much as possible and effectively reduce computation complexity, deep AE is selected as the basic architecture for feature learning in this paper. Then, assuming normal data and outliers may have different neuron behaviors, feature extraction based on deep AE's neuron activation status is implemented on complex and high-dimensional data to generate data for the OOD study.

2) A novel statistical measure for OOD detection is proposed based on local conditional probability (LCP) and data reconstruction. Considering the advantages of different types of OOD methods, e.g., the kernel-function-based density having advantages at addressing data without parametric probability density distribution (PDF) function, the effectiveness of LOF and $k$NN on using neighbor information, this paper proposed a new metric combining these two ideas. The new metric considers calculating probability via neighbors' kernel distance, using the local conditional probability to reconstruct data, and describing OOD data via reconstruction error. It can achieve comparative superiority over conventional statistical measures in OOD detection. Details are presented in Section 3. 

3) The proposed method performs better than conventional OOD detection methods. This paper selects four conventional statistical measures for OOD study, such as KD, LOF, MD, and $k$NN. Then, experiments based on image benchmark datasets and an industrial dataset are implemented. Results illustrate the proposed method's effectiveness and superiority on OOD data description detection.

The rest of this paper is organized as follows. Section II introduces how to describe the OOD data and formulate the detection of OOD data. Section III presents the methodology of the proposed method, including the framework description and LCP calculation. Section IV implemented experiments on image benchmark datasets. Discussion based on numerical results is done. Finally, Section V concluded and outlooked this paper.

\section{Description and Related work of OOD detection}
\begin{figure}
\centering
\includegraphics[scale=0.6]{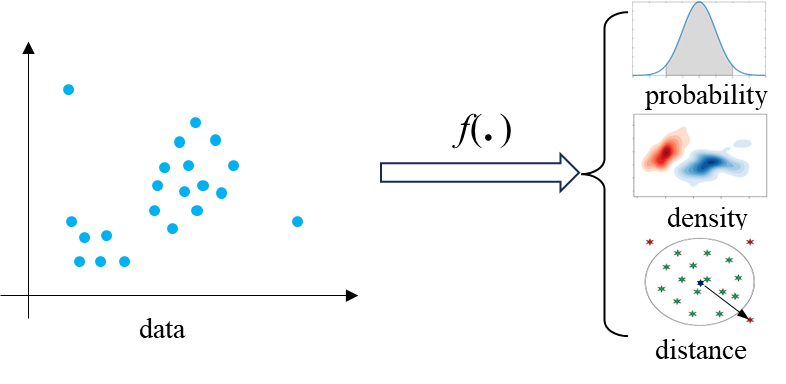}
\caption{Description of OOD data detection}
\label{f1}
\end{figure}

According to the idea of OOD detection methods with different statistical measures above, we can formulate the process of detecting OOD as Fig. \ref{f1}.

Fig. \ref{f1} shows that OOD data are usually distant from the distribution of normal data and have a low density. Based on this assumption, we can easily leverage some data structural characteristics, like distribution probability, density, or distance, to design rules distinguishing normal and outlier data. Then, the formula of OOD detection can be expressed as:
\begin{equation}
score(x) =SM (f(x), f(D_T))
\label{e1}
\end{equation}
where $\textit{score}(\cdot)$ is defined as the numeric calculation of describing a data based on a specific structural characteristic mentioned above. $f(\cdot)$ is the function for feature learning, e.g., kernel learning in OC-SVM and SVDD \cite{b30}, deep feature learning, etc. Thus, representative features can replace the original data for better data description and OOD study. The function $SM(\cdot)$ is the statistical measure of the select structural characteristic, which usually calculate the difference or similarity between testing data $x$ and the training dataset $D_T$, e.g., KD, MD, $k$NN, and LOF.

1) Kernel Density (KD)

KD estimation is developed to address the problem of nonparametric density estimation. When the natural distribution of data is hard to be described by an existing parametric probability density distribution function, the kernel function was applied for density estimation \cite{b23}, as the following equation.
\begin{equation}
score(x_t)=\frac{1}{\left|D_T\right|} \sum_{x_i\in D_T} k(x_t, x_i)
\end{equation}
where, $k(\cdot,\cdot)$ is a kernel function, usually chosen as a Gaussian kernel. The kernel density (i.e., OOD score) is calculated through the average kernel distance between the testing data $x_t$ and the whole training data $D_T$. Then, based on the idea that outliers have a lower density than normal data, we can detect OOD data based on low score values. 

2) Mahalanobis Distance (MD)

With the assumption of Gaussian distributed data, MD is verified as effective in defining the OOD score for anomaly detection. For example, assuming the output of the classification model in each class is approximate to be Gaussian, MD is used and verified superior in adversarial data detection \cite{b15}. The calculation of MD score is calculated below.
\begin{equation}
score(x_t)=(x_t-{\mu})^{\top} {\Sigma}^{-1}(x_t-{\mu}) \\
\end{equation}
\begin{equation}
 \left\lbrace
\begin{aligned}
&\mu=\frac{1}{|D_T|} \sum_{x_i\in D_T} x_i\\
&{\Sigma}=\frac{1}{|D_T|} \sum_{x_i\in D_T}(x_i-{\mu})(x_i-{\mu})^{\top}\\
\end{aligned}
\right.
\end{equation}

Based on the score of MD measurement, OOD data is assumed to have a larger distance than normal data to the center of the given data distribution.

3) $k$-Nearest Neighbors ($k$NN)

The idea of $k$NN in outlier detection is similar to that of KD and MD. It mainly calculates the $k$NN distance and uses it for the OOD score directly, as defined below
\begin{equation}
score(x_t)=\frac{1}{k} \sum_{x_i\in N_k(x_t)} \|x_t-x_i\|^2
\end{equation}
where, $N_k(x_t)$ represents the $k$-nearest neighbors of $x_t$. It is seen that $k$NN utilizes the Euclidean distance instead of the kernel distance in KD. Moreover, $k$NN considers the local distance as the OOD score instead of the global distance in KD and MD. Similarly, if the score of a testing data point is high, the data is assumed as OOD data.

4) Local Outlier Factor (LOF) \cite{b24}

 LOF also makes use of the local characteristic instead of the global characteristic. It describes the relative density of the testing data with respect to its neighborhood. Its calculation is first to find $k$-nearest neighbors of the testing data $x_t$, then to compute the local reachability densities of $x_t$ and all its neighbors. The final score is the average density ratio of $x_t$ to its neighbors, as expressed below.
\begin{equation}
score(x_t)=\frac{1}{\left|N_k(x_t)\right|} \sum_{x_i \in N_k(x_t)} \frac{lrd(x_i)}{lrd(x_t)}
\end{equation}

\begin{equation}
lrd(x)=\frac{\left|N_k(x)\right|}{\sum_{x_i \in N_k(x)} d(x, x_i)}
\end{equation}

where, $lrd(\cdot)$ is the local reachability density function, $d$ is the distance function.  Then, under the assumption that OOD data are drawn from a different distribution than normal data, so OOD data have larger local reachability density, namely higher LOF scores.

Then, based on the scores calculated via different statistical measures, the rule for determining normal and outlier data can be expressed as a Boolean function, as below.
\begin{equation}
\delta(score, \theta)=\left\lbrace 
\begin{aligned}
&1,\ \text{normal data};\\ 
&0,\ \text{outlier};\\
\end{aligned}  \right.
\end{equation}
where,  $\theta$ is a given threshold for OOD detection.

\section{Methodology }
\begin{figure}
\centering
\includegraphics[scale=.5]{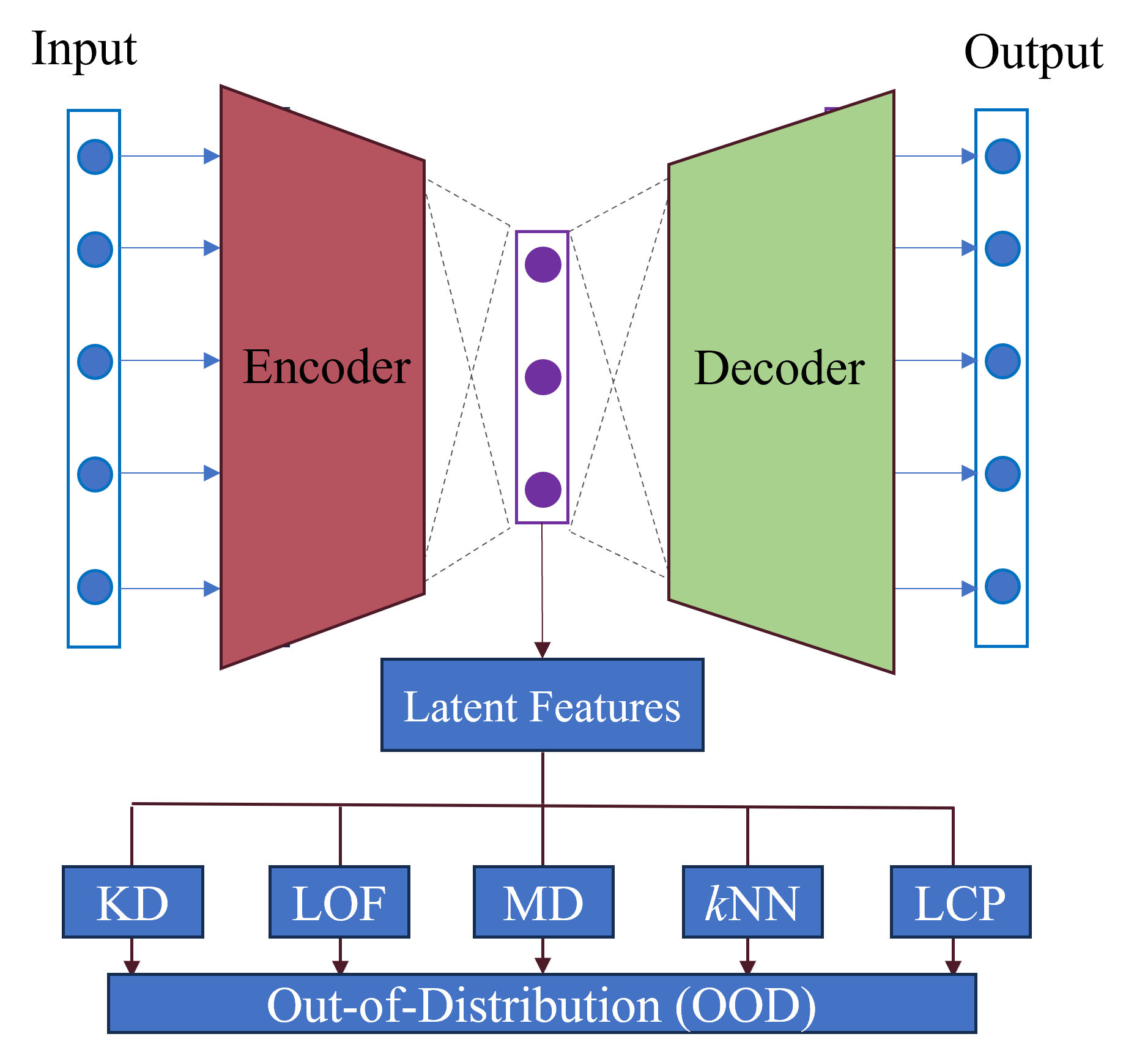}
\caption{The framework of the proposed OOD detection method}
\label{f2}
\end{figure}

From the above description in Section II, the core idea of OOD detection is to use statistical measure, e.g., distribution probability or structural characteristics (distance and density), to distinguish normal and outlier data. However, in terms of complex and high-dimensional data, data's distribution and structural characteristics are not as apparent as that of low-dimensional data. In this case, data transformation is required to simplify data. Therefore, this paper proposes a general framework of the OOD detection method via statistical measures in Fig. \ref{f2}. 

Fig. \ref{f2} shows the framework of the proposed OOD detection method, which makes use of statistical measures as the score function in (\ref{e1}), then determines outlier or normal data. Considering the structural characteristics of high-dimensional data are not readily computed, so a deep AE architecture is used for feature learning and dimensionality reduction first, as shown in Fig. \ref{f2}. Then, some conventional statistical measures are used to calculate the OOD score based on this framework. In this paper, a new statistical measure for better OOD description and detection performance is also come up with, namely the LCP measure in Fig. \ref{f2}. Details for its implementation are described as follows.

\subsection{activation traces}
As mentioned, the proposed framework utilizes deep AE for feature learning and dimensionality reduction. Considering normal and outlier data may have different neuron behaviours in deep learning, this paper proposes to use latent features' neuron activation status for better OOD study instead of original data. This idea is inspired by \cite{b20}, which proposed that using neurons’ status with respect to a testing data point is a good way to describe data’s behavior. For example, the neuron’s activation value is one of the most straightforward descriptions of data behaviors with respect to the given DL model, as well as the sign of neuron outputs. Then, based on these activation statuses, some neuron coverage metrics \cite{b21} were defined to describe the data’s behaviors. In this paper, we can simply define the activation values as the trace of neuron behavior.

By assuming a given deep learning architecture consisting of a set of neurons $N = \{n_1, n_2, \cdots\}$, then, for a given testing data $x \in \{X|x_1, x_2, \cdots\}$, its activation trace with respect to $N $can be defined as $A(x)=\sigma_N(x)$. Generally, considering a DL model contains a huge number of neurons, it is more effective to calculate the activation trace of an ordered (sub)set of neurons $N_s\subseteq N$, instead of all neurons. Inspired by the paper \cite{b22}, which demonstrates abnormal and normal data have different neuron activation statuses on the information channel and the activation network, this paper considers choosing the most active neurons in testing in order to reduce feature dimensionality and computation cost. Moreover, according to the description of the proposed framework, an AE architecture is used for feature representation from high-dimensional and complex data space to low-dimensional space. AEs have the advantage of learning the most representative features with low rank so that they can preserve important information of original data in the dimensionality reduction process. Therefore, the activation traces of active neurons in the latent space of AE are calculated as the features for OOD detection in accordance with the proposed framework in Fig. \ref{f2}. 
%This method has a lower dimensionality for computation reduction and a good ability to distinguish normal and abnormal data behaviors, so the input features $x$ in above methods will be replaced by $A(x)$ in experiments of this paper.

\subsection{local conditional probability (LCP)}
Then, based on the data’s features of activation trace,  different statistical measures are applied to calculate the OOD data score, like KD, MD, $k$NN, and LOF in Fig. \ref{f2}.  
However, summarizing these measures, it is found that the main ideas of these methods are to leverage the average distance as the representation of probability or density in OOD scoring. For example, KD uses the average kernel distances from the global dataset, MD to calculate the Mahanalobis distance. Unlike KD and MD, $k$NN calculates the average Euclidean distance in a local neighborhood domain. Even though LOF is totally different from the other methods, it also makes use of local neighbors’ information, namely the average reachability in the OOD score definition. With consideration of effectiveness of these methods in OOD detection, this paper propose to combine both neighborhood information and distance to develop a statistical measurement for OOD scoring. 

Firstly, instead of calculating the average distance, this paper proposes to reconstruct the testing data via a linear combination of the given training data and then calculate the reconstruction distance as the final OOD score, as defined below
\begin{equation}
score(x_t)=\|x_t-w_i\cdot x_i\|^2, x_i\in \mathbf{X}
\end{equation}
where $w_i$ is the weight for each individual data point $x_i$. It is known that this reconstruction idea is also a useful strategy in OOD detection and shares some similar concepts with that of the distance-based method. For example, KD and $k$NN can also be regarded to calculate the distance between the testing data $x_t$ and the center of the whole data and neighbors, respectively. The $w_i$ can be thought of as equal weights in KD/$k$NN even though they are not for reconstruction. However, with consideration of data reconstruction, the weights $w_i$ should be specified to well reproduce the testing data $x_t$ and satisfy $ \sum_{} w_i=1$. Therefore, a kind of similarity based on local conditional probability (LCP) is defined as the weights in this paper, as below
\begin{equation}
w_i=\frac{d(x_t,x_i)}{\sum_{0\leq j\leq|D_T|} d(x_t,x_j)}
\end{equation}
where, $d(x_t,x_i)$ is the distance between $x_t$ and $x_i$. With the assumption of data obeying Gaussian distribution, the distance function is usually selected as the Gaussian kernel function $k(x,y)=exp(-\frac{\|x-y\|^2}{2\sigma^2})$. In this way, data can be regarded to distribute as a Gaussian distribution centering on data $x_t$ with a variance $\sigma$, so that it is easy to distinguish data nearby or distant to data $x_t$.

Secondly, as the above description, some distant points have no significant influence on data $x_t$'s reconstruction, it is not necessary to consider all data points in $D_T$ in computation. Therefore, for the consideration of efficient computation, we can consider a neighborhood for data reconstruction. Then, then final ODD score with consideration of LCP can be rewritten as
\begin{equation}
\begin{aligned}
score(x_t)=\|x_t-w_i\cdot x_i\|^2,\\
w_i=\frac{\|x_t-x_i\|^2/2\sigma_i^2}{\sum_{x_j\in N_k(x_t)}\|x_t-x_j\|^2/2\sigma_j^2}\\
\end{aligned}
\end{equation}
Moreover, considering the distances between two points vary a lot and have different effects on the Gaussian kernel, so the value of $\sigma$ in this paper is considered to be selected based on the binary search method in \cite{b25}. 

\section{Experiments and Results}
In this paper, to evaluate the performance of the proposed method on OOD data detection, especially detecting abnormal data for AI quality assurance, two image benchmark datasets, such as MNIST \cite{b2} and CIFAR10 \cite{b26} datasets, and one industrial dataset are studied in this section. Moreover, considering OOD data are out of the studied problem domain, we can simply use an external dataset as outlier for evaluation experiment in this paper, e.g., CIFAR10 data can be anomalies out of the problem domain of MNIST data.

1) Example 1: MNIST dataset

\begin{figure}
\centering
\includegraphics[scale=.5]{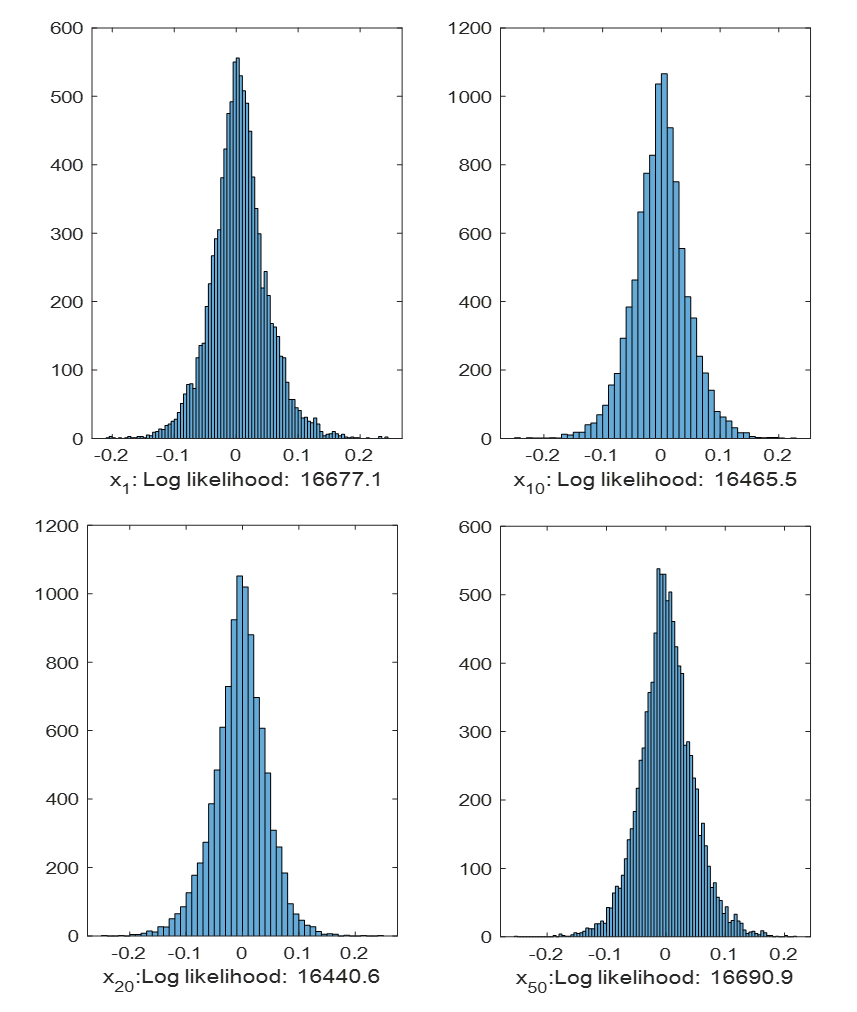}
\caption{Distribution of reconstruction error by using LCP}
\label{ferr}
\end{figure}

The MNIST dataset is a benchmark dataset consisting of images of hard-writing digital numbers from 0 to 9, it is widely used in deep learning studies. It contains 50,000 training images and 10,000 testing images, with a pixel resolution of 28*28. It is seen that the original data is high-dimensional and not very suitable for OOD detection directly. According to the proposed OOD detection framework in Section II, a deep AE with 5 layers is first constructed for feature learning on MNIST data and then trained on the given training dataset. For convenience, we can then extract the activation trace of the output of the encoder as the features in the OOD study since they have acceptable dimensionality and good representation ability to the original data. Subsequently, we can utilize the extracted features for the calculation of OOD scores via different statistical measurements defined in Section III. Based on the proposed LCP measurement, which is actually based on data reconstruction, so first we can show the distribution of reconstruction error by using LCP in data reconstruction, as shown in Fig. \ref{ferr}.

\begin{figure}
\begin{subfigure}{.5\textwidth}
\centering
\includegraphics[scale=.5]{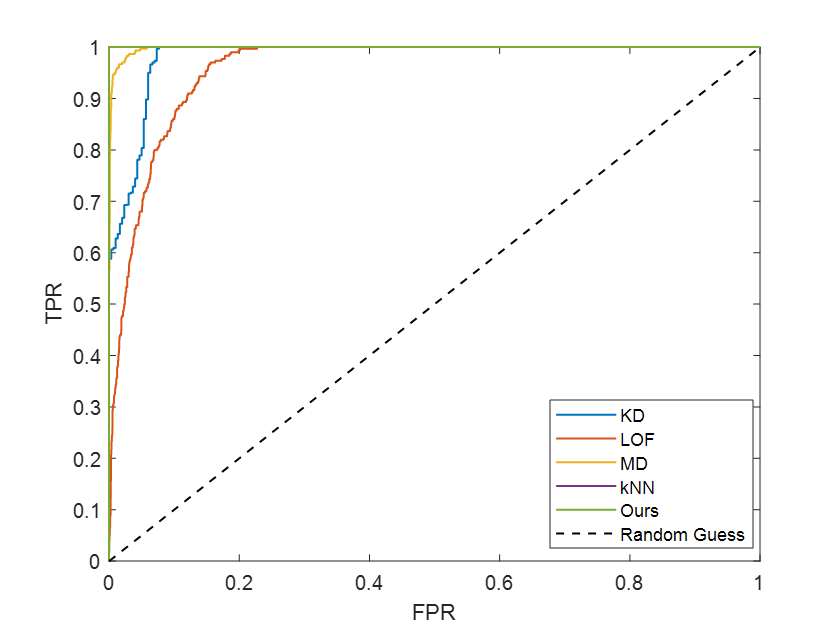}
\caption{noise outliers}
\end{subfigure}
\begin{subfigure}{.5\textwidth}
\centering
\includegraphics[scale=.5]{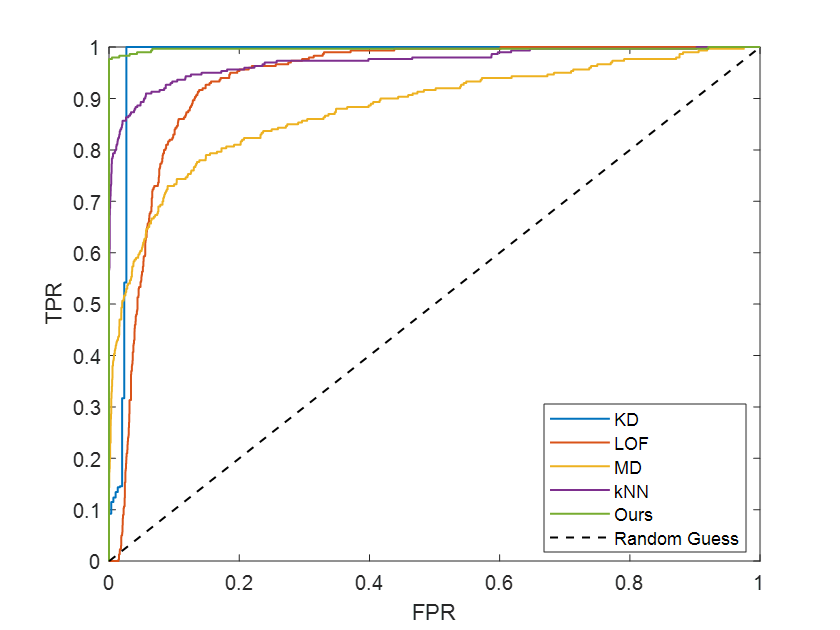}
\caption{CIFAR10 outliers}
\end{subfigure}
\caption{ROC curves of different OOD data detection}
\label{froc}
\end{figure}

Fig. \ref{ferr} plots reconstruction errors in 4 dimensions. It is seen that conditional probability distributions of reconstruction error match well with the Gaussian distribution, which satisfies our hypothesis in Section III.C. Therefore, it is verified feasible to use LCP measurement for OOD data description. Then, we can calculate the reconstruction error via (10) for OOD data detection. In order to generate OOD data for evaluation, random noise and formatted CIFAR10 images are separately considered in the experiments. Results via ROC \cite{b27} curves are shown in Fig. \ref{froc}.

To quantitatively analyze the above results, the following table calculates and presents the AUC values of ROC curves.
\begin{table}[H]
\begin{center}
\caption{AUC-ROC of OOD detection on MNIST dataset}\label{tb1}
\begin{tabular}{cccccc}
\hline
 	&KD	&LOF	 &MD	 &\textit{k}NN	&OURS\\
\hline
O(Noise) &0.9822	&0.94157	&0.9581	&0.9980	&1.0000\\
O(CIFAR) & 0.9787	&0.9328	&0.8794	&0.9713	&0.9962\\
\hline
\end{tabular}
\end{center}
\end{table}

From results of Fig. \ref{froc} and Table \ref{tb1}, we can see that, in comparison among conventional methods, $k$NN and KD perform better than LOF and MD. One possible reason for this is that MD requires strict Gaussian distribution on all data. In this paper's experiments, we did not consider sub-pattern learning in MNIST data, so the latent features destroy this Gaussian assumption on all data. However, $k$NN and LCP can get rid of this global assumption since they use only local information. Moreover, since the proposed LCP inherits the advantages of both KD and $k$NN, it is found to outperform all other methods on noise and CIFAR10 outlier detection from the above results, demonstrating its effectiveness on OOD detection.
\begin{figure}
\centering
\includegraphics[scale=.5]{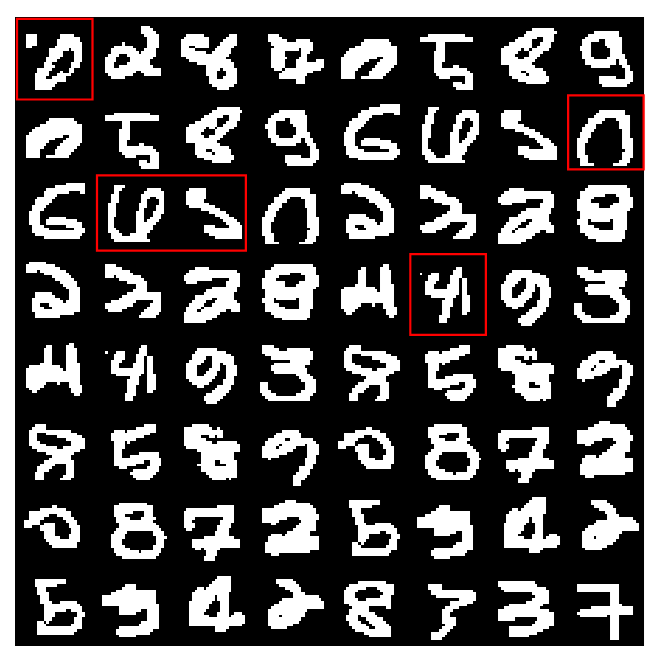}
\caption{MNIST samples with high OOD scores via LCP}
\label{f5}
\end{figure}
Additionally, with the demonstration of LCP's superiority in OOD detection, Fig. \ref{f5} shows some examples of MNIST data with high OOD scores using LCP measurement. These data with high OOD scores are assumed to be corner case data \cite{b28} with a high risk of wrong decision. For example, as seen in Fig. \ref{f5}, some abnormal images with additional patterns, incomplete shape, unsuitable rotation, or shape confusion are found. Therefore, it can further qualitatively demonstrate the ability of LCP measurement in OOD description.

2) Example 2: CIFAR10 dataset

Besides the MNIST dataset, CIFAR10 is also a benchmark dataset in deep learning study, which contains 10 classes of 3*32*32 images. Following the same implementation steps above, we first construct AE for feature learning, then use the learned features for OOD study. To generate OOD data for evaluation, random noise images and reformatted MNIST data are considered outliers in this experiment. Then, OOD detection performance based on the mentioned five statistical measurements is calculated and presented below.
\begin{table}
\begin{center}
\caption{AUC-ROC of OOD detection on CIFAR10 dataset}\label{tb2}
\begin{tabular}{cccccc}
\hline
 	&KD	&LOF	 &MD	 &\textit{k}NN	&OURS\\
\hline
O(Noise) &	0.9878	&0.9906	&0.9858&	0.9788	&0.9880\\
O(MNIST)	&0.9984	&0.7013	&0.9012&0.9789&	1.0000\\
\hline
\end{tabular}
\end{center}
\end{table}
Based on the results of Table \ref{tb2}, we can find the same conclusion that the proposed LCP measurement outperforms other metrics on both noise and MNIST outlier detection. Therefore, the feasibility and effectiveness of the LCP-based OOD detection method are further proved on the CIFAR10 dataset. Moreover, we can also utilize the OOD score via LCP measurement to show some corner case data, as below

\begin{figure}
\centering
\includegraphics[scale=0.5]{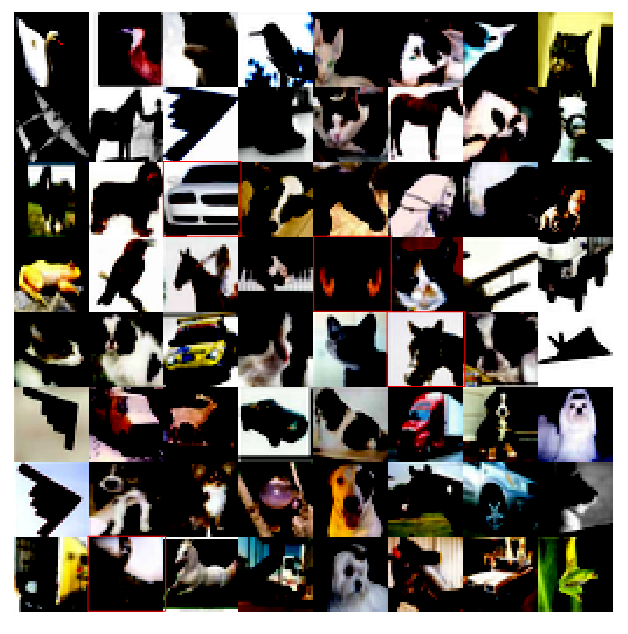}
\caption{CIFAR10 data with high OOD scores via LCP}
\label{f6}
\end{figure}
From the given CIFAR10 images with high OOD scores, we can find that some images are cropped, e.g., images of cat, horse, and car highlighted with red frame in Fig. \ref{f6}. Same to the incomplete images in the MNIST dataset, cropped CIFAR10 images have a loss of information compared with normal data, so they have a risk in training and testing. Through the proposed OOD detection method, it is possible for developers to clean these kinds of data for data quality assurance. 

3) Example 3: German Traffic Sign Recognition Benchmark (GTSRB) dataset \cite{b29}

Furthermore, to study the generation ability of the proposed method, this paper proposes to use an industrial dataset for evaluation, e.g., the GTSRB dataset. This dataset contains 43 classes of traffic signs. By unifying the input images as the size of 3*32*32 and feeding them to train an AE model for feature learning. Then, based on the proposed framework in Fig. \ref{f2}, activation traces of the latent features in AE are used for the OOD study. Four traditional OOD methods based on KD, MD, $k$NN, and LOF are implemented along with the proposed LCP-based method. Numerical results on AUC-ROC are presented in Table \ref{tb3}.

\begin{figure}
\centering
\includegraphics[scale=1]{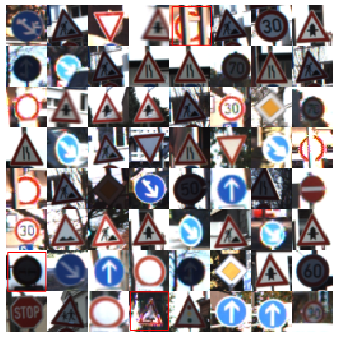}
\caption{GTSRB data with high OOD scores via LCP}
\label{f7}
\end{figure}

\begin{table}
\begin{center}
\caption{AUC-ROC of OOD detection on GTSRB dataset}
\begin{tabular}{cccccc}
\hline
 	&KD	&LOF	 &MD	 &\textit{k}NN	&OURS\\
\hline
O(Noise) &	0.7020	&0.7492	&0.7128	&0.6052	&0.7703\\
O(MNIST)	&0.7175	&0.9593	&0.9790	&0.9729	&0.9985\\
O(CIFAR)	&	0.6263	&0.7874	&0.6244	&0.8620	&0.9147\\
\hline
\end{tabular}
\label{tb3}
\end{center}
\end{table}
In Table \ref{tb3}, outliers based on noise, MNIST data, and CIFAR10 data are implemented, respectively. Through the comparison among different OOD methods, the superiority of the proposed LCP-based method is again verified in the industrial dataset. Moreover, these methods seem to perform better in detecting MNIST outliers than the other two, and this is because MNIST data have more discriminative features compared with GTSRB data. Furthermore, by using the LCP-based data reconstruction error as the OOD data score, the possible outliers inside the original dataset are shown in Fig. \ref{f7}.
Among these GTSRB images with high OOD scores in Fig. \ref{f7}, it is seen that there are some images with low brightness, blurred, or shielded with other objects, which might be the reasons leading to high OOD scores and lower data quality in AIQM.

\section{Conclusions}
In this paper, we addressed the issue of out-of-distribution (OOD) data in AI quality management by proposing a framework that combines deep learning and statistical measures for OOD detection. Initially, we leveraged the strong feature representation and dimensionality reduction capabilities of AE to extract activation traces from hidden neurons as input features for OOD analysis. Subsequently, we employed five statistical measures, namely KD, LOF, MD, $k$NN, and the proposed LCP, using these representative features for OOD detection. Our findings revealed that the proposed LCP, which incorporates both neighbor information and data reconstruction error, outperforms the other measures in OOD detection. Thus, it was proved to be valuable for describing and detecting OOD data. Furthermore, this research extracted reasonable corner case data with high OOD scores from the given datasets, such as MNIST, CIFAR10, and GTSRB datasets. These corner case data have high OOD scores and exhibit abnormal characteristics compared to normal data. Therefore, detecting such data using the proposed method is helpful in future AI quality assurance, particularly for quality analysis related to data security.

\section{Acknowledgement}
This research is supported by the New Energy and Industrial Technology Development Organization (NEDO) project ’JPNP20006’, and JSPS Grant-in-Aid for Early-Career Scientists (Grant Number 22K17961).

\end{document}